\pdfoutput=1
\documentclass{llncs}
\usepackage{llncsdoc}
\usepackage{url}
\usepackage{graphicx}
\usepackage{amsmath}
\usepackage{amssymb}
\usepackage[caption=false]{subfig}
\usepackage{hyperref}
\usepackage{color,colortbl}
\usepackage{footmisc}
\usepackage{hyperref}
\usepackage{eucal}
\usepackage{capt-of}
\usepackage{tabu}
\usepackage{booktabs}

\usepackage{color,soul}
\usepackage[colorinlistoftodos]{todonotes}
\usepackage{xspace}

\newcommand{\citet}[1]{\citeauthor{#1} \shortcite{#1}}

\begin{document}

\title{Text-to-Image Synthesis Based on Machine Generated Captions}

\author{Marco Menardi \inst{1}, Alex Falcon \inst{1}, Saida S.Mohamed \inst{1}, Lorenzo Seidenari \inst{2}, Giuseppe Serra \inst{1}, Alberto Del Bimbo \inst{2} \and Carlo Tasso \inst{1}}
\institute{Artificial Intelligence Laboratory \\
University of Udine, Italy\\ \email{\{menardi.marco, falcon.alex, mahmoud.saidasaadmohamed\}@spes.uniud.it
 \{giuseppe.serra, carlo.tasso\}@uniud.it} \\
 \and Media Integration and Communication Center \\
University of Firenze, Italy\\
\email{\{lorenzo.seidenari, alberto.delbimbo\}@unifi.it}}
\maketitle
\begin{abstract}
Text-to-Image Synthesis refers to the process of automatic generation of a photo-realistic image starting from a given text and is revolutionizing many real-world applications. In order to perform such process it is necessary to exploit datasets containing captioned images, meaning that each image is associated with one (or more) captions describing it. Despite the abundance of uncaptioned images datasets, the number of captioned datasets is limited. 
To address this issue, in this paper we propose an approach capable of generating images starting from a given text using conditional GANs trained on uncaptioned images dataset. In particular, uncaptioned images are fed to an Image Captioning Module to generate the descriptions. Then, the GAN Module is trained on both the input image and the ``machine-generated'' caption. 
To evaluate the results, the performance of our solution is compared with the results obtained by the unconditional GAN.
For the experiments, we chose to use the uncaptioned dataset LSUN-bedroom.
The results obtained in our study are preliminary but still promising.
\end{abstract}
\section{Introduction}
Text-to-Image Synthesis, also called Conditional Image Generation, is a process that consists in generating a photo-realistic image given a textual description. It is a challenging task and it is revolutionizing many real-world applications. For example, starting from a Digital Library of adventure books it could be possible to enrich the reading experience with computer-generated images of the locations explored in the story, while a Digital Library of recipe books may be enriched with images representing the steps involved in a given recipe. In addition, such images may be used to exploit Information Retrieval systems based on visual similarity. Due to its great potentiality and usefulness, it raised a lot of interest in the research fields of Computer Vision, Natural Language Processing, and Digital Libraries. 

One of the main approaches used for the text-to-image task involves the use of  Generative Adversarial Networks (GAN) \cite{GANs}: starting from a given textual description, GANs can be conditioned on text \cite{Reed2016}, \cite{Reed2016LearningDraw}, \cite{StackGAN++} in order generate high-quality images that are highly related to the text meaning. 

To condition a GAN on text, captioned images datasets are needed, meaning that one (or more) captions must be associated to each image. Despite the large amount of uncaptioned images datasets, the number of captioned datasets is limited. For example, the LSUN-bedroom dataset contains $\sim3,000,000$ images \cite{LSUN}, but it does not contain the associated captions. This may lead to a difficulty in training a conditional GAN to generate bedroom images related to a given textual description, such as ``a bedroom with blue walls, white furniture and a large bed''. 

\begin{figure}
\centering
  \includegraphics[width=0.7\linewidth]{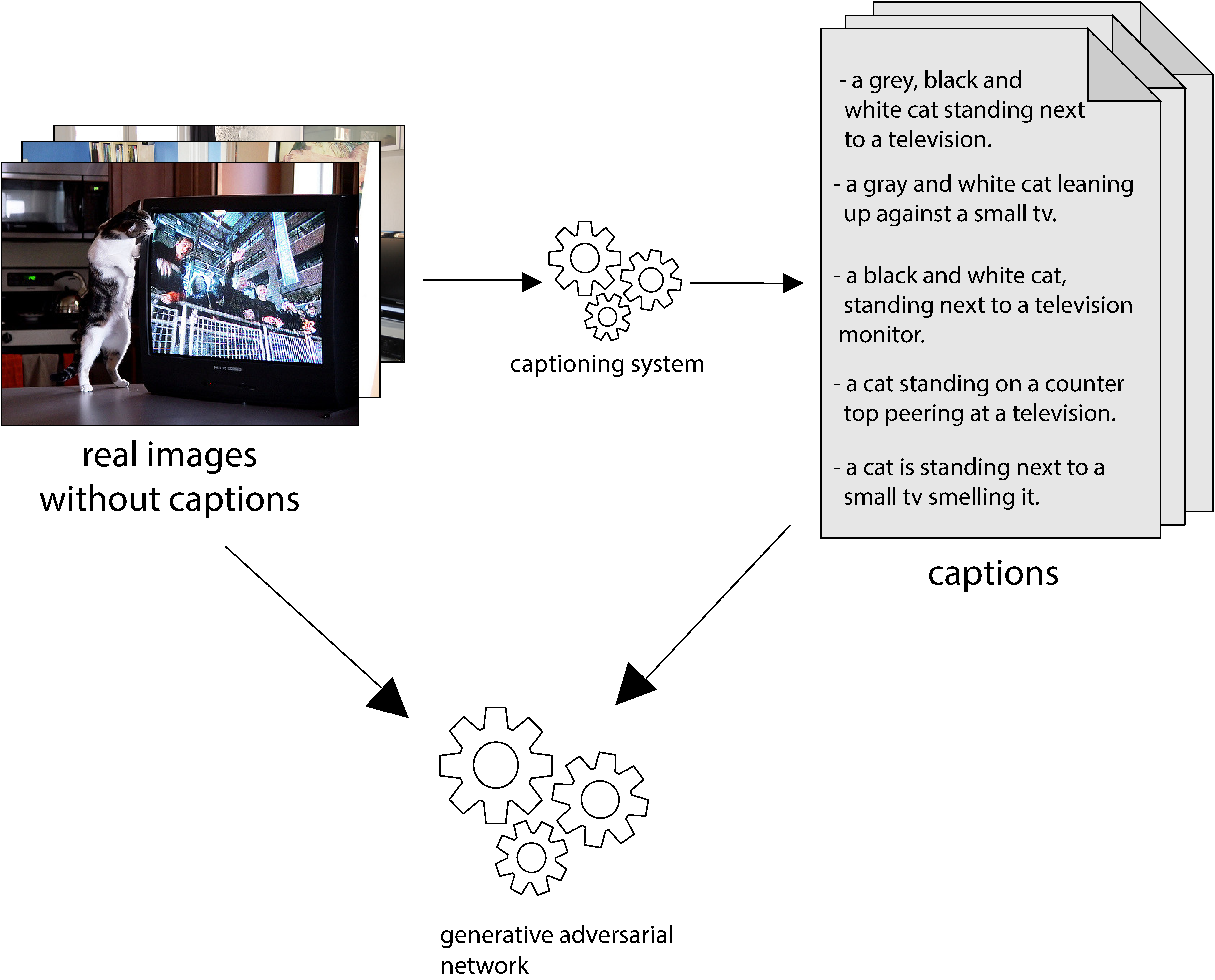}
  \caption{Pipeline: images are fed to a captioning system that outputs its captions. The generated captions and the images are then given as input for training the conditional GAN.}
  \label{pipeline_illustrator_3}
\end{figure}

In this paper we propose an innovative, though quite simple approach to address this issue. First of all, a captioning system (that we call Image Captioning Module) is trained on a generic captioned dataset and used to generate a caption for the uncaptioned images. Then, the conditional GAN (that we call GAN Module) is trained on both the input image and the ``machine-generated'' caption. A high-level representation of the architecture is shown in Figure \ref{pipeline_illustrator_3}. To evaluate the results, the performance of the GAN using ``machine-generated'' captions are compared with the results obtained by the unconditional GAN. To test and evaluate our pipeline, we are using the LSUN-bedroom \cite{LSUN} dataset.

The results obtained in the experiments are very preliminary yet very promising. According to our study, the GAN Module does not learn how to produce meaningful images, with respect to the caption meaning, and we hypothesize that this is due to the ``machine-generated'' captions we use to condition the GAN Module. The Image Captioning module is trained on the COCO dataset \cite{MSCOCO}, which contains captioned images for many different classes of objects and intuitively this should lead the Image Captioning Module to learn how to produce captions for bedroom images as well. Despite being able to produce the desired captions, we notice that the ``machine-generated'' captions are often too similar and not detailed for different bedroom images. The last section of the paper proposes some approaches that can deal with these problems.

\section{Related Work}
In 2014, Goodfellow et al.\ introduced Generative Adversarial Networks (GAN) \cite{GANs}, a  generative model framework that consists in training simultaneously two models: a generator network and a discriminator one. The generator network has the task of generating images as real as possible, while the discriminator network has to distinguish the generated images from the real ones. Generative models are trained to implicitly capture the statistical distribution of training data; once trained, they can synthesize novel data samples, which can be used in the tasks of semantic image editing \cite{generative-visual}, data augmentation \cite{data-augmentation} and style transfer \cite{CycleGAN}. 

GANs can be trained to sample from a given data distribution, in such case a random vector is provided as input to the generator. Otherwise, as in the case of text-to-image synthesis, they can be trained conditionally, meaning that an additional variable is provided as input to control the generator output. In certain formulations the discriminator observes the conditioning variable too, during training. In the literature, several possibilities were tested for the variables used to condition a GAN: attributes or class labels (e.g.\ \cite{Chen2016}, \cite{Odena2016}), images (e.g.\ for the tasks of photo editing \cite{generative-visual} and domain transfer \cite{Isola2016}).

Several methods have been developed to generate images conditioned on text. Mansimov et al.\ \cite{Mansimov2015} built an AlignDRAW model trained to learn the correspondence between text and generated images. Reed et al.\ in \cite{Reed2017} used PixelCNN to generate images using both text descriptions and object location constraints. Nguyen et al.\ \cite{Nguyen2016} used an approximate Langevin sampling approach to generate images conditioned on text, but it required an inefficient iterative optimization process. In \cite{Reed2016}, Reed et al.\ successfully generated $64 \times 64$ images for birds and flowers conditioning on text descriptions. In their follow-up work \cite{Reed2016LearningDraw}, they were able to generate $128 \times 128$ images by using additional annotations on object part locations. Denton et al.\ in \cite{Denton2015} proposed the Laplacian pyramid framework (LAPGANs), which is composed of a series of GANs. A residual image is conditioned at each level of the pyramid on the image of the previous stage to produce an image for the next stage. Also in \cite{Karras2017}, Kerras et al.\ use a similar approach by incrementally adding more layers in the generator and in the discriminator. \cite{StackGAN} and \cite{StackGAN++} suggest the use of a so-called sketch-refinement process, where the images are first generated at low resolutions using a GAN conditioned over the textual description, and then refined with another GAN conditioned on both the image generated at the previous step and the input textual description. \cite{hong2018} and \cite{li2019} infer a semantic label map by predicting bounding boxes and object shapes from the text, and then synthesize an image conditioned on the layout and the text description. A recent work by Qiao et al.\ \cite{qiao2019} uses a three-step approach where it first computes word- and sentence-level embedding from the given textual description, then it uses the embeddings to generate images in a cascaded architecture, and finally starting from the image generated at the previous step it tries to regenerate the original textual description, in order to semantically align with it. Although several different state-of-the-art architectures may be chosen for the task, such as HDGAN \cite{HDGAN}, AttGAN \cite{AttGAN} and BigGAN \cite{BigGAN}, in our pipeline we decided to use StackGAN-v2 \cite{StackGAN++} as the conditional GAN component, given the availability of its code on GitHub.

Recently, several impressive results \cite{ren2017}, \cite{Zhang2017RL}, \cite{self-critical} were obtained for the Image Captioning (or image-to-text) task, which deals with the generation of a caption describing the given image and the objects taking part to it. It is an important task that raises a lot of interest in the Computer Vision and Natural Language Processing research fields. A recent and comprehensive survey about the task is provided by Hossain et al.\ in \cite{captioning_survey}. Some of the approaches used for this task involve the use of Encoder/Decoder networks and Reinforcement learning techniques.

The encoder/decoder paradigm for machine translation using recurrent neural networks (RNNs) \cite{encoder-decoder-RNN} inspired \cite{Karpathy2017}, \cite{show-and-tell} to use a deep convolutional neural network to encode the input image, and a Long Short-Term Memory (LSTM) \cite{long-short-memory} RNN decoder to generate the output caption. 
Given the unavailability of labeled data, recent approaches to the image captioning task involve the use of reinforcement learning and unsupervised learning-based techniques. \cite{ren2017} and \cite{Zhang2017RL} use actor-critic reinforcement learning methods, where a ``policy network'' (the actor) is trained to predict the next word based on the current state, whereas a ``value network'' (the critic) is trained to estimate the reward of each generated word. These techniques overcome the need to sample from the policy (actors) action space, which can be enormous, at the expense of estimating future rewards. 
Another approach, used by Ranzato et al.\ in \cite{teacher-forcing}, consists in applying the REINFORCE algorithm \cite{statistical-gradient}. A limitation of this algorithm consists in the requirement of a context-dependent normalization to tackle  the high variance encountered when using mini-batches.
The approach we are following uses Self-Critical Sequence Training (SCST) \cite{self-critical} which is a REINFORCE algorithm that utilizes the output of its own test-time inference algorithm to normalize the rewards it experiences: doing so, it does not need neither to estimate the reward signal nor the normalization. 

\section{Our Approach} 
\begin{figure}[t]
  \centering
  \includegraphics[width=0.9\linewidth]{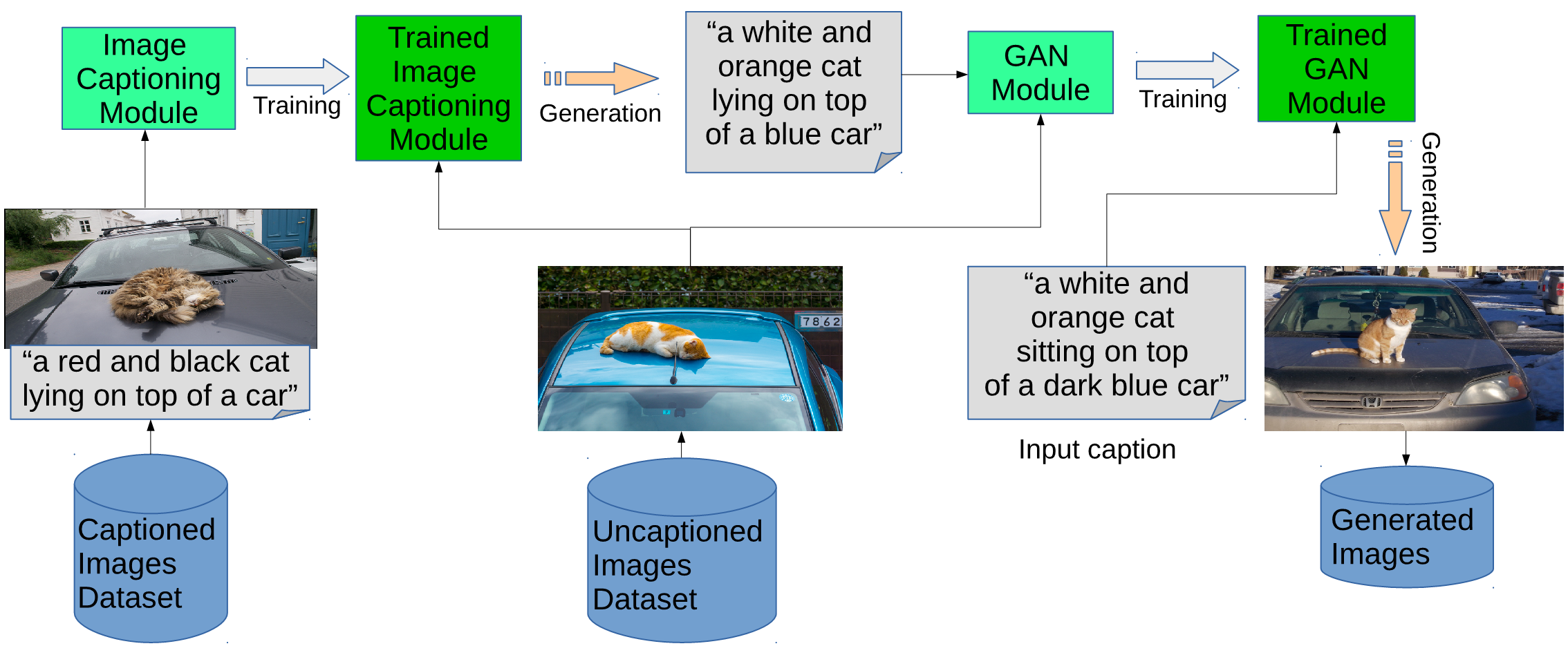} 
  \caption{Our pipeline: captioned images are used to train the Image Captioning Module; uncaptioned images are then captioned through the Trained Image Captioning Module and both the image and the generated captions are used to train the GAN Module; finally, the Trained GAN Module is used to generate an image based on an input caption.}
  \label{pipeline} 
\end{figure}
We propose a pipeline whose goal is to generate images by conditioning on ``machine-generated'' captions. This is fundamental when image captions are not available for a specific domain of interest. Thus, the proposed solution involves the use of a generic captioned dataset, such as the COCO dataset, to make the Image Captioning Module capable of generating captions for a specific domain.

To do so, we want to explore the possibility of using an automatic system to generate textual captions for the images and use them for the training of a Generative Adversarial Network.
For achieving our goal, we built a pipeline composed by an Image Captioning Module and a GAN Module, as shown in Figure \ref{pipeline}. First of all, the Image Captioning Module is trained over a generic captioned dataset to generate multiple captions for the input image. Then, real images are given as input to the Trained Image Captioning Module, which outputs multiple captions for each image. The generated captions together with the images are then fed to the GAN Module, which learns to generate images conditioned on the ``machine-generated'' captions. By feeding the GAN with multiple captions for each image, the GAN can better learn the correspondence between images and captions. 

In the following sections, 
we detail the two modules used in our pipeline: the Image Captioning Module and the GAN Module.
\subsection{Image Captioning Module} 
The goal of the Image Captioning Module is to generate a natural language description of an image. Good performance in this task are obtained by learning a model which is able to first understand the scene described in the image, the objects taking part to it and the relationships between them, and then to compose a natual language sentence describing the whole picture. Given the complexity of such a task, it is still an open challenge in the fields of Natural Language Processing and Computer Vision.

In our pipeline, we are implementing our Image Captioning Module in a similar way as the one proposed in \cite{self-critical}, meaning that we also use a captioning system based on FC models, which is then optimized through Self-Critical Sequence Training (SCST). 

Typical deep learning models used for the Image Captioning task are trained with the ``teacher-forcing'' technique, which consists in maximizing the likelihood of the next ground-truth word given the previous ground-truth word. This has been shown to generate some mismatches between the training and the inference phase, knows as ``exposure bias''. Moreover, the metrics used during the testing phase are non-differentiable (such as BLEU and CIDEr), meaning that the captioning model can not be trained to directly optimize them. To overcome these problems, Reinforcement Learning techniques such as the REINFORCE algorithm have been used. SCST is a variation and an improvement of the popular REINFORCE algorithm that, rather than estimating a baseline to normalize the rewards and reduce variance, utilizes the output of its own test-time inference algorithm to normalize the rewards it experiences. This means that it is forced to improve the performance of the model under the inference algorithm used at test time. Practically, SCST has much lower variance than REINFORCE and can be more effectively trained on mini-batches of samples using SGD. Moreover, it has been shown that SCST system has achieved state-of-the-art performance by optimizing their system using the test metrics of the MSCOCO task.

\subsection{GAN Module} 
The GAN Module has the major role of learning to generate images by conditioning on the ``machine-generated'' captions. In particular, we are using StackGAN-v2 \cite{StackGAN++} as our GAN Module.


StackGAN-v2 consists of a multiple stage generation process, where high-resolution images are obtained by initially generating low-resolution images which are then refined in multiple steps. It consists in a single end-to-end network composed by multiple generators and discriminators in a tree-like structure. Different branches of the tree generate images of different resolutions: at branch $i$, the generator $G_i$ learns the image distribution $p_{G_i}$ at that scale, while the discriminator $D_i$ estimates the probability of a sample being real.\\
Since we are more interested in the conditional case, we are not reporting the loss function used by the generator and the discriminator in the unconditional setting, for which more details can be found in \cite{StackGAN++}.\\
The discriminator $D_i$ takes a real image $x_i$ or a fake sample $s_i$ as input and is trained to classify them as real or fake by minimizing the cross entropy loss:
\begin{equation}
  \label{loss_d}
  \begin{aligned} 
  \mathcal{L}_{D_i} &= \underbrace{-\mathbb{E}_{x_i}\sim p_{data_i}[logD_i(x_i)] - \mathbb{E}_{x_i} \sim p_{G_i}[log(1-D_i(s_i))] + }_{\textrm{unconditional loss}} \\
  &+ \underbrace{-\mathbb{E}_{x_i}\sim p_{data_i}[logD_i(x_i,c)] - \mathbb{E}_{x_i} \sim p_{G_i}[log(1-D_i(s_i,c))]}_{\textrm{conditional loss}}
  \end{aligned}
\end{equation}
where $x_i$ is an image from the true image distribution $p_{data_i}$ at the $i^{th}$ scale, $s_i$ is from the model distribution $p_{G_i}$ at the same scale. 
While StackGAN-v2 \cite{StackGAN++} follows the approach of Reed et al.\  \cite{learning-deep-representations} to pre-train a text encoder to extract visually-discriminative text embeddings of the given description, in our case we use Skip-Thought \cite{skip-thought}, that works at the sentence level, to generate the text embeddings ($c$ in the equations \ref{loss_d} and \ref{joint_gen_loss}). Sentences that share semantic and syntactic properties are mapped to corresponding vector representations \cite{skip-thought}.

The multiple discriminators are trained in parallel each one for a different scale, while the generator is instead optimized to jointly approximate multi-scale image distributions $p_{data_0}, p_{data_1}, ..., p_{data_{m-1}}$ by minimizing the following loss function:
\begin{equation}
  \label{joint_gen_loss}
  \mathcal{L}_G = \sum_{i=1}^m \mathcal{L}_{G_i}, \quad \mathcal{L}_{G_i} = \underbrace{- \mathbb{E}_{s_i}\sim p_{G_i}[logD_i(s_i)]}_{\textrm{unconditional loss}} + \underbrace{- \mathbb{E}_{s_i}\sim p_{G_i}[logD_i(s_i,c)]}_{\textrm{conditional loss}}
\end{equation}
where $L_{G_i}$ is the loss function for approximating the image distribution at the $i^{th}$ scale.
The unconditional loss is used to determine whether the image is real or fake, while the conditional loss is used to determine if the image and the condition match.

\section{Experimental Results}
In this section, we present the preliminary results of the experiments involving the proposed pipeline. The Image Captioning Module was trained on the COCO dataset \cite{MSCOCO}, which contains $120,000$ generic images tagged with categories and captioned by five different sentences each. 

The uncaptioned dataset that we considered is the LSUN \cite{LSUN} dataset, which consists in around one million labeled images for each of the 10 scene categories and 20 object categories \cite{LSUN}. From the LSUN dataset, we first select the $\sim3,000,000$ images tagged with the ``bedroom'' scene category and from that set a subset of $120,000$ images is selected: $80,000$ are then used to train the GAN and $40,000$ as test set. Later on in this paper, the selection of the $\sim3,000,000$ images tagged with the ``bedroom'' scene category is called ``LSUN-bedroom''.

A typical metric used to evaluate both the quality and the diversity of generated images is the Inception Score \cite{Salimans2016}. Unfortunaly, the type of image of the LSUN dataset is very different from those used by ImageNet, therefore it has be shown that the Inception Score is not a good indicator for the quality of generated images \cite{StackGAN++}. So we decided not to report the obtained scores. 

\begin{figure}[h]
  \centering
  \includegraphics[width=0.8\linewidth]{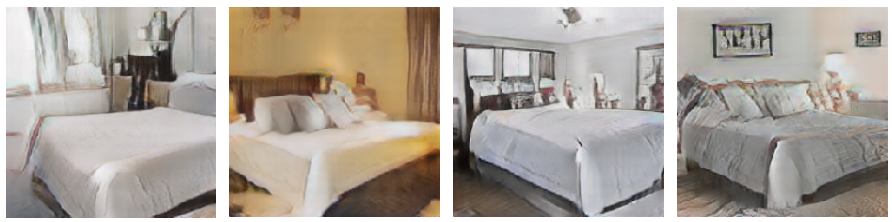}
  \caption{Examples of images generated by the GAN Module trained on the whole LSUN-bedroom dataset.}
  \label{bedroom_all} 
\end{figure}

We performed three experiments over the considered dataset.
The first experiment consists in training the GAN Module on the whole LSUN-bedroom dataset ($\sim3,000,000$ images). This is done because of two reasons: first, it serves as a baseline for the next experiment; second, we compare the results obtained by our computing facilities with the results obtained in \cite{StackGAN++}, since with our graphics card we are limited to a lower batch size of 16. Figure \ref{bedroom_all} shows some examples of generated images, and it is possible to see that the quality of the generated images is similar to those reported in \cite{StackGAN++}. 

\begin{figure}[h]
    \centering
    \includegraphics[width=0.8\linewidth]{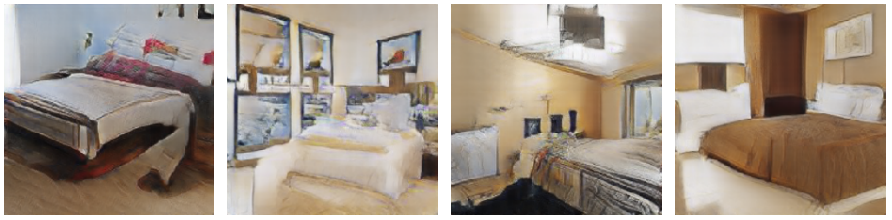}
    \caption{Examples of images generated by the GAN Module trained on a part of the LSUN-bedroom dataset.}
  \label{bedroom_split} 
\end{figure}

Because of computational issues, we decided to explore and understand how the GAN Module performs with less training images. In the second experiment, the training of the GAN Module without conditioning is done on a subset of LSUN-bedroom, consisting of $120,000$ images. Some of the results obtained in this experiment are showed in Figure \ref{bedroom_split}. Although the quality of the generated images is slightly reduced, it is possible to see that the semantic content is still clear and defined.

\begin{figure}
  \centering
  \includegraphics[width=0.9\linewidth]{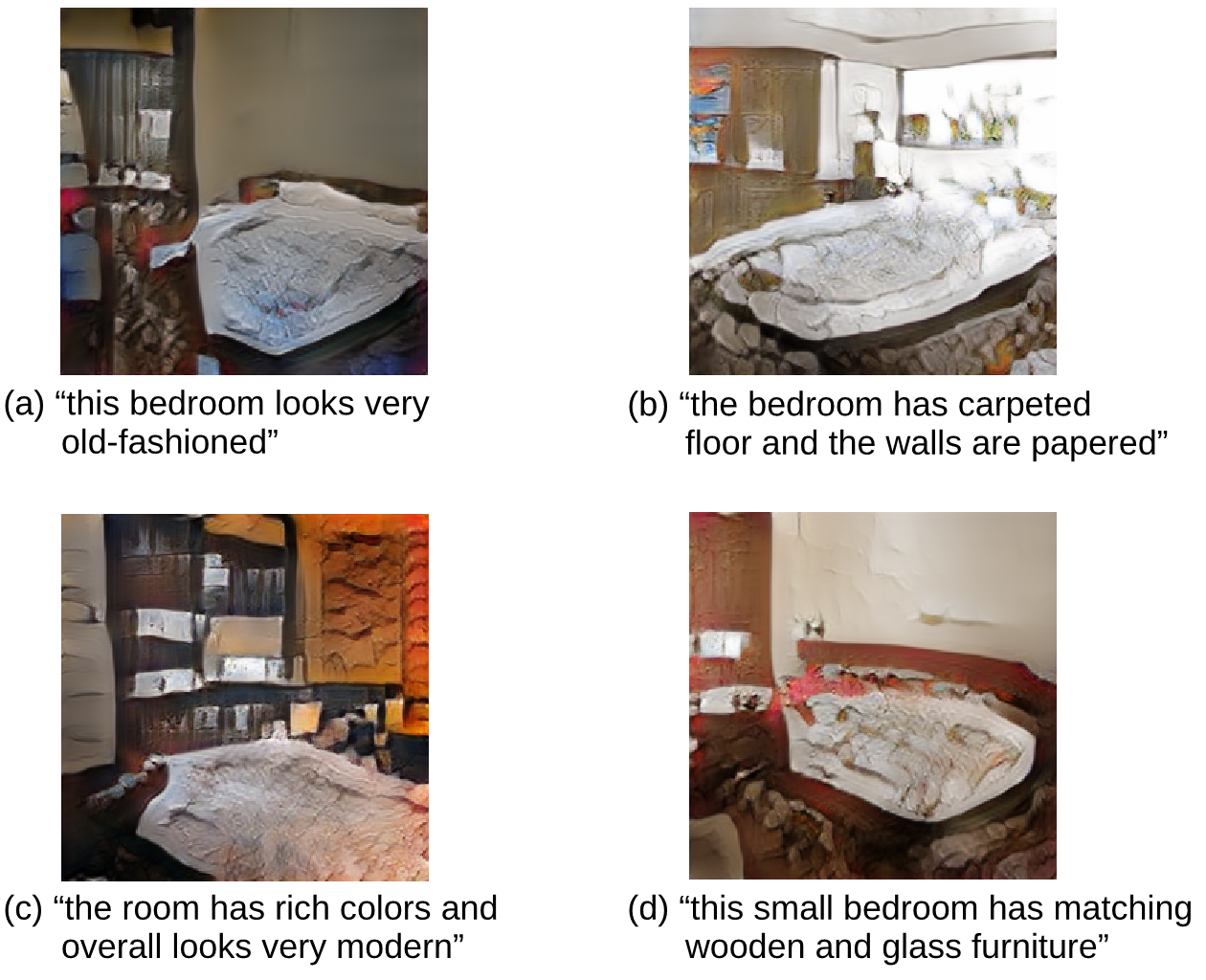}
  \caption{Examples of images generated by the GAN Module trained on a part of the LSUN-bedroom dataset and conditioned on ``machine-generated'' captions.} 
 \label{bedroom_split_captions} 
\end{figure}

Finally, to test our pipeline, we used the Image Captioning Module to generate captions for the images contained in the considered subset of the LSUN-bedroom dataset. Then, the GAN Module was trained on  these same images and conditioned by the ``machine-generated'' captions. About the preliminary results that we obtained, some examples are shown in Figure \ref{bedroom_split_captions}. After a few epochs, the generator produces images that start to look more and more like Fig.\ (b). We suspect the problem is due to the similarity of the ``machine-generated'' captions: the LSUN-bedroom dataset does not come with captions and thus the Image Captioning Module is trained on a generic dataset (COCO) and not for that specific dataset. Because of this, the Image Captioning Module is unable to produce detailed and varied captions for different bedroom images. Then, the captions are used to yield the embeddings, which are also used as noise by the generator. The fact that the noise is almost always the same could be the cause of the observed problem.

\section{Conclusion}
We explored the problem of conditional image generation using Generative Adversarial Networks with machine-generated captions. For this task, we built a pipeline to first generate captions for uncaptioned datasets and then to use the ``machine-generated'' captions to condition a GAN. To test our pipeline, we run experiments on the LSUN-bedroom dataset, which is a subset of the LSUN dataset containing uncaptioned images of bedrooms, and then compare the generated images in the unconditional setting and in the conditional setting where ``machine-generated'' captions are used.

The results observed in the experiments do not achieve success in the task of conditioning with ``machine-generated'' captions. So we identify, analyze, and propose possible solutions to the obstacles that need to be overcome.

The Image Captioning Module that we trained on the COCO dataset seems to generate captions too similar to each other. This is probably related to the fact that more diverse and detailed captions are needed during training in order to achieve significant improvements. During a subsequent review of works on captioning, we found a work from Shetty et al.\ \cite{Shetty2017}, that promises to generate more different captions, instead of variations of the same caption. This result is achieved by using GANs for image captioning instead of other traditional methods. An open question is whether with a bigger dataset the GAN could learn the image-captions correspondence, even when captions are very similar for each image.


An hybrid approach could make our proposed method work by making humans write captions on a subset of the dataset, then use the obtained captions to train a captioning system. For generating human captions, crowdsourcing platforms like Amazon Mechanical Turk could be used.
\bibliographystyle{splncs03}
\bibliography{aiia2017}
\end{document}